%% file: main.tex
\pdfoutput=1  %
\documentclass[letterpaper, 10 pt, conference]{ieeeconf}  %

\IEEEoverridecommandlockouts  %
\input{preamble}

\title{\LARGE \bf
\methodname:\\ Joint Garment Identification and State Estimation under Self-Occlusion
}

\author{Jens Lundell$^{1,*}$, Alberta Longhini$^{2,*}$, Lei Shu$^{2}$, Hemal Arora$^{2}$, Evan Robert$^{2}$,\\ Carlota Parés-Morlans$^{2}$, Marcel Büsching$^{3}$, Mårten Björkman$^{3}$ and Jeannette Bohg$^{2}$%
\thanks{$^{*}$Equal contribution; the first two authors share first authorship.}%
\thanks{$^{1}$University of Turku, Finland. {\tt\small jens.lundell@utu.fi}}%
\thanks{$^{2}$Stanford University, USA.}%
\thanks{$^{3}$KTH Royal Institute of Technology, Sweden.}%
}

\begin{document}

\bstctlcite{IEEEcompactrefs}

\maketitle
\thispagestyle{plain}
\pagestyle{plain}

\begin{abstract}

Garment manipulation requires inferring both garment identity and complete configuration from partial, self-occluded observations. This creates a chicken-and-egg problem: completing a partial observation requires a garment-specific deformation prior, while
  identifying the garment requires a configuration that explains the observation. We introduce ClothAtlas, which jointly identifies garments and estimates their complete configurations from partial point clouds and a library of canonical meshes. A
  shared graph-conditioned flow-matching model samples configurations for each candidate mesh and ranks candidates by observation fit. The model accommodates different mesh shapes, vertex counts, and connectivities, allowing new garments to be added
  from canonical meshes alone without retraining. On held-out DexGarmentLab garments, ClothAtlas ranks the correct mesh first among 221 candidates for all 22 test garments from a single flat observation. From a single self-occluded observation, it
  identifies 15/22 garments, compared with 8/22 for a deterministic graph estimator. Persistent vertex identities support grasp-target localization on real garments, reducing mean error from 15.4 to 6.4\,cm overall and from 17.0 to 8.3\,cm on occluded
  targets compared with dense descriptor matching. Project page: \mbox{\url{https://clothatlas.github.io/}}

\end{abstract}
\acresetall %

\suppressfloats[t]
\begin{figure}[!t]
    \centering
    \input{Images/tikz/fig1_overview}
    \caption{(a) \methodname{} combines a canonical mesh library and partial
    depth observations to identify the garment and estimate its complete
    configuration. Red points show the observations while the blue surface shows
    the estimated configuration, including regions occluded from the camera.
    (b) Four reconstructed configurations illustrate persistent point tracking
    through deformation and occlusion. The camera icon indicates the observation
    viewpoint and the colored tracks connect the estimated positions of three
    mesh vertices through the lifting action.}
    \label{fig:teaser}
\vspace{-1pt}
\end{figure}
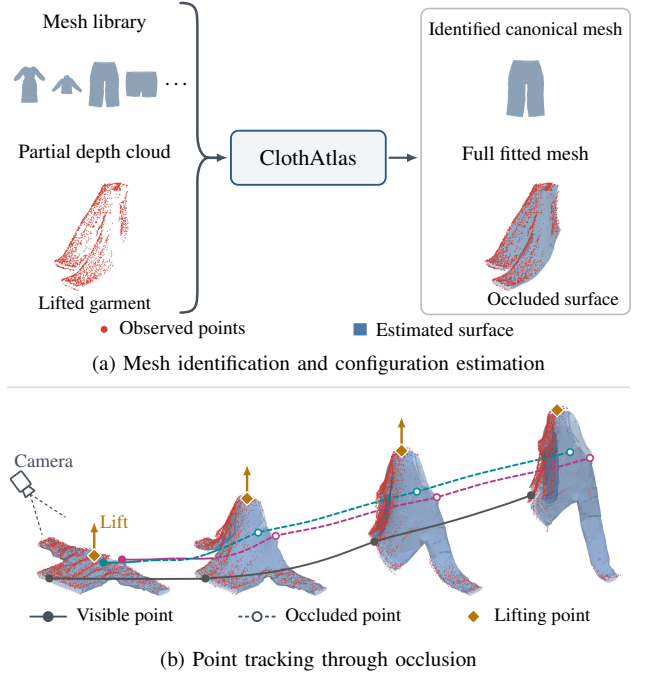

\input{sections/01_Introduction}

\input{sections/02_RW}

\input{sections/03_P_Formulation}
\input{sections/04_Method}

\input{sections/05_Experiments}

\input{sections/06_Conclusion}

\section*{Acknowledgments}
This work was supported in part by Agile Robotics.
Alberta Longhini was supported by a Wallenberg--Bienenstock Postdoctoral Fellowship.
The authors wish to acknowledge CSC -- IT Center for Science, Finland, for generous computational resources.
The authors used ChatGPT (OpenAI) to assist with figure preparation and code development.

\bibliographystyle{IEEEtran}
\bibliography{references}

\end{document}

%% file: preamble.tex
\usepackage[OT1]{fontenc}
\usepackage{graphicx}
\usepackage[caption=false,font=footnotesize]{subfig}  %
\usepackage{amsmath}
\usepackage{amssymb}
\usepackage{booktabs}
\usepackage{cite}
\usepackage[dvipsnames]{xcolor}
\usepackage[hidelinks]{hyperref}
\usepackage{pgfplots}
\usepackage{multirow}
\pgfplotsset{compat=1.18}
\usepgfplotslibrary{statistics}
\usetikzlibrary{arrows.meta}

\usepackage{xspace}
\newcommand{\methodname}{ClothAtlas\xspace}

\definecolor{lavender}{rgb}{0.9, 0.9, 0.98}
\definecolor{orange}{RGB}{252, 130, 62}
\definecolor{darkgreen}{RGB}{0, 180,0}
\definecolor{lightgray}{gray}{0.9}
\definecolor{lightblue}{rgb}{0.68,0.85,0.9}
\definecolor{lightorange}{rgb}{1.0, 0.8, 0.6}
\definecolor{myblue}{RGB}{30, 100, 200}
\definecolor{figblue}{HTML}{2A78D6}
\definecolor{figorange}{HTML}{EB6834}
\definecolor{figgreen}{HTML}{1BAF7A}
\definecolor{figred}{HTML}{D03B3B}

\newcommand{\remark}[3]{}

\newcommand{\todo}[1]{}

\newcommand{\equationref}[1]{\hyperref[#1]{(\ref*{#1})}}
\newcommand{\figref}[2][]{\hyperref[#2]{Figure~\ref*{#2}\if\relax\detokenize{#1}\relax\else(#1)\fi}}
\newcommand{\tabref}[1]{\hyperref[#1]{Table~\ref*{#1}}}
\newcommand{\secref}[1]{\hyperref[#1]{Sec.~\ref*{#1}}}
\newcommand{\algoref}[1]{\hyperref[#1]{Alg.~\ref*{#1}}}

\newcommand{\defref}[1]{\hyperref[#1]{Definition~\ref*{#1}}}

\makeatletter
\long\def\@makecaption#1#2{%
\ifx\@captype\@IEEEtablestring
  \footnotesize
  \setbox\@tempboxa\hbox{\normalfont #1:\ }%
  \noindent\unhbox\@tempboxa#2\par
  \@IEEEtablecaptionsepspace
\else
  \@IEEEfigurecaptionsepspace
  \setbox\@tempboxa\hbox{\footnotesize #1.~~ #2}%
  \ifdim \wd\@tempboxa >\hsize
    \setbox\@tempboxa\hbox{\footnotesize #1.~~ }%
    \parbox[t]{\hsize}{\footnotesize \noindent\unhbox\@tempboxa#2}%
  \else
    \ifcenterfigcaptions \hbox to\hsize{\footnotesize\hfil\box\@tempboxa\hfil}%
    \else \hbox to\hsize{\footnotesize\box\@tempboxa\hfil}%
    \fi
  \fi
\fi}
\makeatother

\usepackage{acronym}
\acrodef{pomdp}[POMDP]{Partially Observable Markov Decision Process}
\acrodef{gnn}[GNN]{graph neural network}
\acrodef{rgb}[RGB]{Red-Green-Blue}
\acrodef{rgbd}[RGB-D]{Red-Green-Blue-Depth}
\acrodef{mlp}[MLP]{Multi-layer Perceptron}
\acrodef{ddpm}[DDPM]{Denoising Diffusion Probabilistic Model}
\acrodef{mse}[MSE]{mean squared error}
\acrodef{nn}[NN]{Neural Network}
\acrodef{gps}[GPS]{General, Powerful, Scalable}
\acrodef{gns}[GNS]{Graph Network Simulator}
\acrodef{map}[MAP]{maximum a posteriori}
\acrodef{rl}[RL]{Reinforcement Learning}
\acrodef{3d}[3D]{three-dimensional}
\acrodef{adalnzero}[AdaLN-Zero]{zero-initialized adaptive layer normalization}
\acrodef{cd}[CD]{Chamfer distance}
\acrodef{ci}[CI]{confidence interval}
\acrodef{fm}[FM]{flow matching}
\acrodef{gpu}[GPU]{graphics processing unit}
\acrodef{pca}[PCA]{principal component analysis}
\acrodef{pck}[PCK]{percentage of correct keypoints}
\acrodef{pcn}[PCN]{Point Completion Network}
\acrodef{rms}[RMS]{root mean square}
\acrodef{vlm}[VLM]{vision-language model}
\acrodef{vla}[VLA]{vision-language-action model}

\acrodef{ra}[RA]{Rigid Alignment}
\acrodef{dgnn}[D-GNN]{Deterministic GNN}
\acrodef{cr}[C+R]{Completion + Registration}
\acrodef{lre}[LRE]{Learned Retrieval Embedding}
\acrodef{sp}[SP]{Static Position}
\acrodef{ddm}[DDM]{Dense Descriptor Matching}

\usepackage[linesnumbered,ruled,vlined]{algorithm2e}
\SetKwInput{KwInput}{Input}
\SetKwInput{KwOutput}{Output}

%% file: Images/tikz/fig1_overview.tex
\begingroup%
\providecommand{\figoneassetdir}{Images/fig1}%
\ifdefined\figoneunit\else\newlength{\figoneunit}\fi%
\setlength{\figoneunit}{\dimexpr\columnwidth*2/7\relax}%
\definecolor{figoneblue}{HTML}{245B87}%
\definecolor{figonegray}{HTML}{46515B}%
\definecolor{figonepoints}{HTML}{D43D32}%
\definecolor{figonemesh}{HTML}{4C78A8}%
\definecolor{figoneborder}{HTML}{B7BDC3}%
\definecolor{figonefill}{HTML}{F1F5F8}%
\definecolor{figonelift}{HTML}{B5770C}%
\begin{tikzpicture}[x=\figoneunit,y=-\figoneunit,
    text=black,font=\fontsize{7.6}{9}\selectfont,
    label/.style={anchor=north,inner sep=0pt,outer sep=0pt,
        text height=5.4pt,text depth=1.5pt},
    heading/.style={label,anchor=north,font=\fontsize{8}{9}\selectfont,
        text height=6pt},
    smalllabel/.style={label,font=\fontsize{7.3}{8.5}\selectfont},
    icon/.style={anchor=north west,inner sep=0pt,outer sep=0pt},
    flow/.style={-{Latex[length=1.5mm,width=1.2mm]},draw=figonegray,
        line width=0.8pt,line cap=round}]

\path[use as bounding box] (0,.22) rectangle (3.5,3.88);

\node[label] at (.55,.305) {Mesh library};
\foreach \k/\x in {0/.11,1/.31,2/.51,3/.71}{
    \node[icon] at (\x,.54)
        {\includegraphics[width=.18\figoneunit,height=.29\figoneunit]{\figoneassetdir/a_library_\k.pdf}};
}
\node[label,anchor=north west,font=\fontsize{8.2}{9}\selectfont] at (.925,.615) {\ldots};
\node[label] at (.55,1.005) {Partial depth cloud};
\node[icon] at (.06,1.15)
    {\includegraphics[width=.98\figoneunit,height=.65\figoneunit]{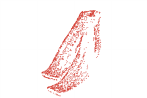}};
\node[smalllabel,font=\fontsize{7.1}{8.5}\selectfont] at (.55,1.815) {Lifted garment};

\draw[figonegray,line width=.95pt,line cap=round,line join=round]
    (1.01,.25) .. controls (1.10,.25) and (1.10,.29) .. (1.10,.39)
    -- (1.10,.95) .. controls (1.10,1.045) and (1.135,1.08) .. (1.16,1.08)
    .. controls (1.135,1.08) and (1.10,1.115) .. (1.10,1.21)
    -- (1.10,1.77) .. controls (1.10,1.87) and (1.10,1.91) .. (1.01,1.91);
\draw[flow,line width=1pt] (1.16,1.08) -- (1.265,1.08);
\draw[draw=figoneblue,fill=figonefill,rounded corners=2.8pt,line width=.7pt]
    (1.28,.935) rectangle (2.11,1.225);
\node[inner sep=0pt,font=\fontsize{8.4}{10}\selectfont]
    at (1.695,1.08) {\methodname};
\draw[flow] (2.13,1.08) -- (2.285,1.08);

\draw[figoneborder,rounded corners=2.8pt,line width=.6pt]
    (2.30,.28) rectangle (3.42,1.92);
\node[label,font=\fontsize{7.1}{8.5}\selectfont] at (2.86,.345) {Identified canonical mesh};
\node[icon] at (2.56,.545)
    {\includegraphics[width=.60\figoneunit,height=.33\figoneunit]{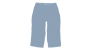}};
\node[label] at (2.86,1.015) {Full fitted mesh};
\node[icon] (fitted) at (2.37,1.15)
    {\includegraphics[width=.98\figoneunit,height=.65\figoneunit]{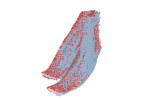}};
\foreach \corner/\anchor in {nw/north west,ne/north east,sw/south west}{
    \coordinate (fitted-\corner) at (fitted.\anchor);
}
\input{\figoneassetdir/surface_annotation.tex}
\node[label,anchor=north east,font=\fontsize{7.1}{8.5}\selectfont]
    (occluded-label) at (3.36,1.795) {Occluded surface};

\fill[figonepoints] (.60,2.00) circle[radius=1.3pt];
\node[anchor=west,inner sep=0pt,font=\fontsize{7.3}{8.5}\selectfont]
    at (.68,2.00) {Observed points};
\draw[draw=figonemesh,fill=figonemesh,line width=.35pt]
    (1.94,1.97) rectangle (2.005,2.03);
\node[anchor=west,inner sep=0pt,font=\fontsize{7.3}{8.5}\selectfont]
    at (2.055,2.00) {Estimated surface};

\node[heading] at (1.75,2.13) {(a) Mesh identification and configuration estimation};
\draw[black!18,line width=.45pt] (.08,2.31) -- (3.42,2.31);

\node[icon] (tracking) at (.08,2.37)
    {\includegraphics[width=3.34\figoneunit,height=1.12\figoneunit,keepaspectratio]{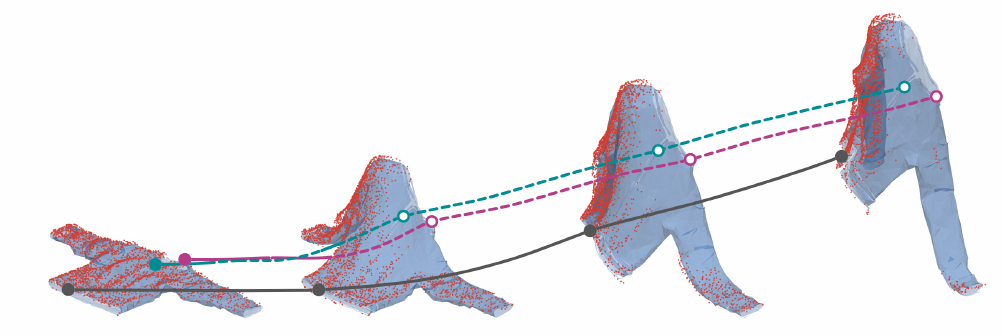}};
\foreach \corner/\anchor in {nw/north west,ne/north east,sw/south west}{
    \coordinate (tracking-\corner) at (tracking.\anchor);
}
\input{\figoneassetdir/tracking_annotations.tex}
\begin{scope}[shift={(tracking-camera)},x=1pt,y=1pt,rotate=\figonetrackingcameraangle]
    \draw[draw=figonegray,line width=.45pt,dash pattern=on 1.2pt off 1pt]
        (5,0) -- (20,-5.5)
        (5,0) -- (20,5.5);
    \draw[draw=figonegray,fill=white,line width=.65pt,rounded corners=.5pt]
        (-4,-2.6) rectangle (2.3,2.6);
    \draw[draw=figonegray,fill=white,line width=.65pt,line join=round]
        (2.3,-1.4) -- (5,-2.1) -- (5,2.1) -- (2.3,1.4) -- cycle;
\end{scope}
\node[anchor=south west,inner sep=0pt,xshift=-3pt,yshift=6pt,
    text=figonegray,font=\fontsize{7.1}{8}\selectfont] at (tracking-camera) {Camera};
\foreach \i in {0,1,2}{
    \draw[draw=figonelift,line width=.9pt,-{Latex[length=1.4mm,width=1.1mm]}]
        (tracking-lift-\i) -- (tracking-lift-end-\i);
}
\foreach \i in {0,1,2,3}{
    \node[rectangle,rotate=45,draw=white,fill=figonelift,line width=.45pt,
        minimum size=3.6pt,inner sep=0pt] at (tracking-lift-\i) {};
}
\node[anchor=west,inner sep=0pt,xshift=2pt,text=figonelift,
    font=\fontsize{7.1}{8}\selectfont] at (tracking-lift-end-0) {Lift};
\draw[figonegray,line width=.7pt] (.20,3.54) -- (.40,3.54);
\filldraw[draw=figonegray,fill=figonegray,line width=.7pt]
    (.30,3.54) circle[radius=1.7pt];
\node[anchor=west,inner sep=0pt,font=\fontsize{7.1}{8}\selectfont]
    at (.45,3.54) {Visible point};
\draw[figonegray,line width=.7pt,dash pattern=on 1.5pt off 1pt]
    (1.32,3.54) -- (1.52,3.54);
\filldraw[draw=figonegray,fill=white,line width=.7pt]
    (1.42,3.54) circle[radius=1.7pt];
\node[anchor=west,inner sep=0pt,font=\fontsize{7.1}{8}\selectfont]
    at (1.57,3.54) {Occluded point};
\node[rectangle,rotate=45,draw=white,fill=figonelift,line width=.45pt,
    minimum size=3.6pt,inner sep=0pt] at (2.58,3.54) {};
\node[anchor=west,inner sep=0pt,font=\fontsize{7.1}{8}\selectfont]
    at (2.69,3.54) {Lifting point};

\node[heading] at (1.75,3.72) {(b) Point tracking through occlusion};

\end{tikzpicture}%
\endgroup%

%% file: sections/01_Introduction.tex
\section{Introduction}
\label{sec:intro}

Garment manipulation tasks are typically specified with respect to the garment rather than the scene: grasp the left shoulder, align the two sleeves, fold along the mid-line. Executing these tasks requires locating task-relevant locations on the garment, including those hidden from the camera view. A complete mesh with persistent vertex identities provides a task-agnostic interface for this purpose: it maintains vertex correspondences through deformation and occlusion, such as in \figref[b]{fig:teaser}, while exposing the complete geometry needed for downstream reasoning about the garment state.  Such a representation can support mesh-based planning~\cite{huang2022mesh,tian2025diffusion} and provide persistent spatial cues to generalist policies~\cite{li2025crayonrobo,li2025hamster,huang2024rekep,chen2026pointingvla}.
Recovering hidden and self-occluded geometry, however, creates a coupled inference problem: completing a partial observation requires a garment-specific deformation prior, while identifying the garment requires finding a configuration that explains the observation~\cite{longhini2025unfolding}. Existing estimators typically assume a predefined garment category~\cite{chi2021garmentnets,xue2023garmenttracking} or a known canonical mesh~\cite{huang2022mesh,tian2025diffusion,huang2023self,longhini2025cloth}. We instead consider the setting in which neither garment identity nor its complete configuration is known, and must be inferred jointly from partial observations.

We introduce \methodname{}, which jointly selects a canonical mesh from a library and estimates its complete deformed configuration from partial point clouds (\figref[a]{fig:teaser}). A shared graph-conditioned \ac{fm} model samples configurations for each candidate independently at each observation, while accumulated reconstruction-fit scores rank candidates over time without a separate recognition model. Trained only on simulated observation--mesh pairs, it accepts meshes with different shapes, vertex counts, and connectivities, allowing unseen candidates to be added from their canonical meshes alone without additional fine-tuning.

In simulation, \methodname{} retrieves the exact mesh for 15/22 unseen garments among 221 candidates from a single occluded observation, compared with 8/22 for a deterministic graph estimator. On real-world garments, it reduces mean grasp-target localization error from  17.0 to 8.3\,cm compared with the best tracking baseline when the target is occluded.

In summary, our contribution is a method for joint garment identification and complete configuration estimation from partial observations, enabled by a shared graph-conditioned \ac{fm} estimator that generalizes to unseen canonical meshes without retraining and ranks library candidates by reconstruction fit across observations.

%% file: sections/02_RW.tex
\section{Related Work}
\label{sec:related}

\textbf{Steering policies with spatial cues.}
Generalist policies map visual observations and language instructions to robot actions and have shown impressive garment manipulation capabilities~\cite{physicalintelligence2025pi05,bjorck2025gr00t, su2026demavla}. Because language specifies {\em where\/} to act only coarsely, recent work conditions action generation on spatial cues, such as points, visual prompts, trajectories, or keypoint constraints \cite{li2025crayonrobo,li2025hamster,huang2024rekep,chen2026pointingvla}. These interfaces are demonstrated primarily on rigid or articulated objects, where a cue's location follows the object's pose and remains recoverable from the current view. For deformable objects this no longer holds: the position of a surface point is a function of the garment's full configuration, and once the point is occluded nothing in the observation determines where it went. \methodname{} targets this gap by estimating a complete mesh with persistent vertex identities. Therefore, a cue identified once can be localized as the garment deforms. The estimate is produced from depth observations alone, independently of the downstream policy and robot embodiment. Section~\ref{sec:exp-real-grasp} evaluates the resulting localization accuracy, while integration with a policy remains future work.

\textbf{Object-centric representations for garment manipulation.}
 Object-centric representations broadly follow two directions.
  Dense correspondence models transfer manipulation targets across garments
  and configurations~\cite{florence2018dense,ganapathi2021dense,wu2024unigarmentmanip,wang2025dexgarmentlab},
  but search only observed surfaces and thus do not directly represent
  occluded targets.
  Complete-state estimators recover unobserved geometry through known-template
  deformation~\cite{wang2023trtm,huang2022mesh,huang2023self,longhini2025cloth},
  category-level canonical representations or completed-point-cloud
  tracking~\cite{chi2021garmentnets,xue2023garmenttracking,zhang2025gapt},
  or generative priors~\cite{li2024guided,guo2024diffusion}.
  Most closely related to our work is UniClothDiff~\cite{tian2025diffusion}, which generates complete configurations from partial point clouds but requires training a separate generative model per mesh. \methodname{} overcomes this issue by training a shared estimator across candidate mesh topologies to produce complete meshes.
  
\textbf{Canonical-mesh identification.}
Retrieval-and-deformation methods for mesh identification jointly select and fit candidates from a finite library~\cite{uy2021joint,di2023ured,di2024shapematcher}, but primarily consider static objects and use dedicated retrieval representations. \methodname{} instead evaluates each candidate through reconstruction fit using
a shared graph-conditioned \ac{fm} estimator, accumulating evidence across observations. %
Moreover, \methodname{} can integrate a new garment from its canonical mesh alone, without candidate-specific demonstrations or retraining.

%% file: sections/03_P_Formulation.tex
\section{Problem Formulation}
\label{sec:problem}

Given a library of $K$ canonical garment meshes
${\mathcal{L}=\{\bar{\mathcal{M}}_1,\ldots,\bar{\mathcal{M}}_K\}}$
and a sequence of partial point-cloud observations
$\mathbf{P}_{1:t}=(\mathbf{P}_1,\ldots,\mathbf{P}_t)$,
our goal is to jointly identify a candidate
$\bar{\mathcal{M}}_c\in\mathcal{L}$ that best represents the garment
and estimate its complete deformed mesh $\mathcal{M}_t^c$
at observation $t$. Although the true canonical mesh may be absent from
$\mathcal{L}$, we assume that a candidate mesh $\bar{\mathcal{M}}_c$ can represent
the garment with sufficient accuracy throughout the observation sequence. Each observation
$\mathbf{P}_t\in\mathbb{R}^{N_t\times3}$ contains segmented points
on the visible garment surface, acquired by a fixed, calibrated
depth camera in a common coordinate frame.

We define a canonical mesh $\bar{\mathcal{M}}_c=(\bar{\mathbf{V}}_c,\mathcal{T}_c)$
as a flat triangular mesh with faces $\mathcal{T}_c$ and reference vertex
positions $\bar{\mathbf{V}}_c\in\mathbb{R}^{N_v^c\times3}$.
Its deformed mesh $\mathcal{M}_t^c=(\mathbf{V}_t^c,\mathcal{T}_c)$ shares
the same connectivity, while its vertex positions
$\mathbf{V}_t^c\in\mathbb{R}^{N_v^c\times3}$ define its configuration
at observation $t$.

We formulate the problem as inferring the joint posterior
$p(c,\mathbf{V}_t^c\mid\mathbf{P}_{1:t},\mathcal{L})$
over the candidate index $c\in\{1,\ldots,K\}$ and its configuration
$\mathbf{V}_t^c$.
Assuming the configuration $\mathbf{V}_t^c$ depends on the library $\mathcal{L}$ only through
$\bar{\mathcal{M}}_c$, the joint posterior factorizes as
\begin{equation}
\begin{aligned}
p(c,\mathbf{V}_t^c\mid\mathbf{P}_{1:t},\mathcal{L})
&=p(c\mid\mathbf{P}_{1:t},\mathcal{L}) p(\mathbf{V}_t^c\mid\bar{\mathcal{M}}_c,\mathbf{P}_{1:t}),
\end{aligned}
\label{eq:factorization}
\end{equation}
where the first factor describes uncertainty over which canonical
  mesh represents the garment, and the second describes uncertainty
  over its complete configuration given that canonical mesh.

%% file: sections/04_Method.tex
\section{Method}
\label{sec:method}

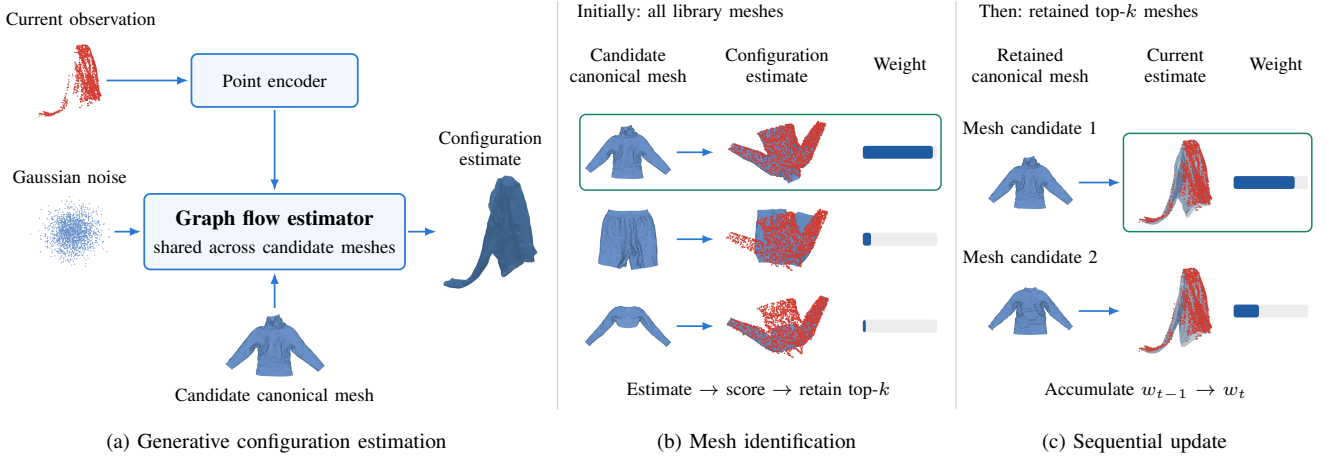
\begin{figure*}[t]
    \centering
    \input{Images/tikz/fig2_overview}
    \caption{\emph{\methodname} overview.
    (a)~A shared graph-conditioned generative model samples complete
    configurations from a partial observation, Gaussian noise, and a
    candidate canonical mesh, which supplies reference vertex positions and
    connectivity. The same model is used for every candidate mesh.
    (b)~We score each candidate by the observation fit of its configuration
    estimate, accumulate evidence across observations, and retain the $k$
    highest-weight candidates after the initial identification period.
    (c)~For each retained candidate, we sample a new configuration from the
    current observation and update its weight using the accumulated evidence.
    We report the highest-weight candidate's configuration.
    }
    \label{fig:method}
\end{figure*}

\methodname{} approximates the joint posterior in
Eq.~\eqref{eq:factorization} by generating complete
garment configurations and comparing their fit to the observations.
For each candidate canonical mesh, a conditional flow-matching
model generates configurations from the current point cloud.
We identify the canonical mesh by accumulating observation-fit
scores in the candidate weights across observations, and report
the highest-weight candidate and its configuration estimate.
\figref{fig:method} provides an overview of the method.

\subsection{Generative Configuration Estimation}
\label{sec:diffusion}
We approximate the configuration factor
${p(\mathbf{V}_t^c\mid\bar{\mathcal{M}}_c,\mathbf{P}_{1:t})}$,
the second factor in Eq.~\eqref{eq:factorization}, with
${q_\phi(\mathbf{V}_t^c\mid\bar{\mathcal{M}}_c,\mathbf{P}_t)}$,
which uses only the candidate canonical mesh and the current point cloud.
We learn $q_\phi$ with a conditional \ac{fm} model~\cite{lipman2023flow} whose architecture builds
upon \mbox{UniClothDiff}~\cite{tian2025diffusion}.
As shown in \figref[a]{fig:method}, flow matching generates a configuration sample by iteratively transforming
Gaussian noise into vertex positions using a learned velocity field
conditioned on the canonical mesh and the observation:
\begin{equation}
    u_\phi:
    \bigl([\mathbf{V}(\sigma) \,\|\, \bar{\mathbf{V}}_c],
    \mathcal{T}_c,\mathbf{e}_t,\sigma\bigr)
    \mapsto\hat{\mathbf{u}}\in\mathbb{R}^{N_v^c\times3},
    \label{eq:flow_velocity}
\end{equation}
where $\mathbf{V}(\sigma)$ are noisy vertex positions, $\sigma$ is the noise
scale, $\bar{\mathcal{M}}_c=(\bar{\mathbf{V}}_c,\mathcal{T}_c)$ is the
candidate canonical mesh, and $\mathbf{e}_t$ is the observation embedding.
Here $\|$ denotes feature concatenation.
The output $\hat{\mathbf{u}}$ predicts the rate of change of
vertex positions with respect to $\sigma$. Integrating this
field yields a complete configuration sample.

We train $u_\phi$ using the conditional flow-matching loss
\begin{equation}
    \mathcal{L}_{\text{\acs{fm}}}
    =\mathbb{E}\Bigl[
        \frac{1}{3N_v^c}
        \bigl\|\hat{\mathbf{u}}-\mathbf{u}^\star\bigr\|^2
    \Bigr],
    \label{eq:flow_loss}
\end{equation}
where $\hat{\mathbf{u}}$ is the velocity predicted by $u_\phi$
and $\mathbf{u}^\star$ is the target velocity.
To construct the target velocity, we sample Gaussian noise
$\boldsymbol{\epsilon}\sim\mathcal{N}(\mathbf{0},\mathbf{I})$
and a noise fraction $\sigma\in(0,1)$, compute noisy vertices
$\mathbf{V}(\sigma)=(1-\sigma)\mathbf{V}_t^c
+\sigma\boldsymbol{\epsilon}$ from the ground-truth
configuration $\mathbf{V}_t^c$, and set
$\mathbf{u}^\star=\boldsymbol{\epsilon}-\mathbf{V}_t^c$.

At inference, we generate a configuration by integrating the
learned velocity field from $\sigma=1$ to $\sigma=0$, starting
from Gaussian noise
$\mathbf{V}_t^c(1)\sim\mathcal{N}(\mathbf{0},\mathbf{I})$.
For integration, we use the explicit Euler update
\begin{equation}
    \mathbf{V}_t^c(\sigma_{j+1})
    =\mathbf{V}_t^c(\sigma_j)-\delta_j
    u_\phi\!\left(
        [\mathbf{V}_t^c(\sigma_j)\,\|\,\bar{\mathbf{V}}_c],
        \mathcal{T}_c,\mathbf{e}_t,\sigma_j
    \right),
    \label{eq:flow_euler}
\end{equation}
where $\delta_j=\sigma_j-\sigma_{j+1}$ is the step size
along the schedule $1=\sigma_0>\cdots>\sigma_J=0$.
The final state $\mathbf{V}_t^c(0)$ is the generated configuration.

During training, the observation embedding
$\mathbf{e}_t=h_\psi(\mathbf{P}_t,\mathbf{P}_{t-\Delta})$
encodes the current point cloud and an earlier point cloud,
providing temporal context for configuration estimation.
However, at inference, we supply
the current cloud to both encoder slots, giving
$\mathbf{e}_t=h_\psi(\mathbf{P}_t,\mathbf{P}_t)$.

We replace the
mesh-specific patch decomposition and interpolation
map from \mbox{UniClothDiff}~\cite{tian2025diffusion} with a \ac{gnn} $u_\phi$. This allows a
single model to process meshes with different vertex counts and
connectivities without retraining. UniClothDiff instead requires
a separate model for each mesh, making evaluation across our
candidate library impractical, which is why we omit it from our
experimental baselines. Further details about $u_\phi$ are given in \secref{sec:exp-setup}.

\subsection{Canonical-Mesh Identification from Observation Fit}
\label{sec:filter}
\label{sec:tracking}

We approximate the candidate posterior
$p(c\mid\mathbf{P}_{1:t},\mathcal{L})$,
the first factor in Eq.~\eqref{eq:factorization}, by accumulating
observation-fit evidence in the candidate weights across observations.
We compare candidate canonical meshes by how well their
estimated configurations fit the observed point clouds
(\figref[b]{fig:method}).
The approximation takes the form
\begin{equation}
    p(c\mid\mathbf{P}_{1:t},\mathcal{L})
    \approx \widehat{p}_t(c)
    =
    \begin{cases}
        w_t(c), & c\in\mathcal{C}_t,\\
        0, & c\notin\mathcal{C}_t,
    \end{cases}
\end{equation}
where $\mathcal{C}_t\subseteq\{1,\ldots,K\}$ contains the
candidate indices retained for evaluation at observation $t$,
and $w_t(c)\geq0$ are weights satisfying
$\sum_{c\in\mathcal{C}_t}w_t(c)=1$.

To update these weights, we evaluate how well each candidate
$c\in\mathcal{C}_t$ explains the current observation $\mathbf{P}_t$.
For each candidate $c$, we generate a configuration estimate $\hat{\mathbf{V}}_t^c$
using Eq.~\eqref{eq:flow_euler}
and measure its observation-fit error using the directional
mean squared point-to-surface distance
\begin{equation}
    d_c(\mathbf{P}_t,\hat{\mathbf{V}}_t^c)
    =\frac{1}{|\mathbf{P}_t|}
    \sum_{\mathbf{p}\in\mathbf{P}_t}
    \min_{\mathbf{x}\in\mathcal{S}(\hat{\mathbf{V}}_t^c,\mathcal{T}_c)}
    \|\mathbf{p}-\mathbf{x}\|_2^2,
    \label{eq:likelihood}
\end{equation}
where $\mathcal{S}(\hat{\mathbf{V}}_t^c,\mathcal{T}_c)$ is the triangulated surface
defined by the vertex positions and candidate connectivity.

We convert this observation-fit error into a score
\begin{equation}
    \ell_t(c)
    =\exp\!\left[
        -\frac{d_c(\mathbf{P}_t,\hat{\mathbf{V}}_t^c)}{\lambda}
    \right],
    \label{eq:fresh_evidence}
\end{equation}
where $\lambda>0$ controls sensitivity to the error.
We use this score to update the weights as
\begin{equation}
    w_t(c)
    =\frac{
        w_{t-1}(c)^\gamma\,\ell_t(c)
    }{
        \sum_{c'\in\mathcal{C}_t}
        w_{t-1}(c')^\gamma\,\ell_t(c')
    },
    \qquad 0<\gamma\leq1,
    \label{eq:evidence}
\end{equation}
where $w_0(c)=1/K$ and $\gamma$ controls the influence of past observations:
$\gamma=1$ retains all accumulated scores, whereas $\gamma<1$
progressively discounts older scores.
This update defines the energy-based approximation $\widehat{p}_t(c)$
to the candidate posterior.

To report a single garment-state estimate,
we select the candidate with the largest updated weight:
\begin{equation}
    \hat{c}_t=\arg\max_{c\in\mathcal{C}_t}w_t(c),
    \label{eq:candidate_readout}
\end{equation}
corresponding to the mesh $\hat{\mathcal{M}}_t=
(\hat{\mathbf{V}}_t^{\hat{c}_t},\mathcal{T}_{\hat{c}_t})$.
This mesh serves as a surrogate when the true canonical
mesh is absent from the library.

For each candidate $c$, we can also use Eq.~\eqref{eq:flow_euler}
to generate $S$ configurations $\mathbf{V}_t^{(c,s)}$,
$s=1,\ldots,S$, using independent noise samples. In this case,
we use the configuration with the smallest observation-fit error
in Eq.~\eqref{eq:likelihood} as $\hat{\mathbf{V}}_t^c$ for both the weight update
and the reported garment-state estimate.

To reduce the cost of evaluating the candidate library, we
prune candidates after an initial identification period.
We evaluate the full library,
$\mathcal{C}_t=\{1,\ldots,K\}$, for $t\leq T_{\mathrm{id}}$.
After the update at $t=T_{\mathrm{id}}$, we retain the $k$
highest-weight candidates for all subsequent observations.
Pruned candidates remain excluded, and Eq.~\eqref{eq:evidence}
normalizes subsequent weights over the retained set.
\figref[c]{fig:method} illustrates configuration estimation and weight updates
for these retained candidates.
\algoref{alg:filter} summarizes the complete procedure,
including multiple-sample selection and candidate pruning.

\begin{algorithm}[!htbp]
\caption{Sequential candidate and configuration estimation.}
\label{alg:filter}
\SetAlgoLined
\KwInput{Observations $\mathbf{P}_{1:T}$; library $\mathcal{L}$; trained models $(h_\psi,u_\phi)$; sampling settings $(S,\{\sigma_j\}_{j=0}^J)$; scoring parameters $(\lambda,\gamma)$; pruning settings $(T_{\mathrm{id}},k)$.}
$\mathcal{C}\gets\{1,\ldots,K\}$; $w_0(c)\gets1/K$ for $c\in\mathcal{C}$\;
\For{$t=1,\ldots,T$}{
    $\mathbf{e}_t\gets h_\psi(\mathbf{P}_t,\mathbf{P}_t)$\;
    \ForEach{$c\in\mathcal{C}$}{
        Draw $\mathbf{V}_t^{(c,s)}$, $s=1,\ldots,S$, using
        Eq.~\eqref{eq:flow_euler}\;
        $\hat{s}_t^c\gets\arg\min_{s\in\{1,\ldots,S\}}
        d_c(\mathbf{P}_t,\mathbf{V}_t^{(c,s)})$\;
        $\hat{\mathbf{V}}_t^c\gets\mathbf{V}_t^{(c,\hat{s}_t^c)}$\;
        Compute $\ell_t(c)$ using Eq.~\eqref{eq:fresh_evidence}\;
    }
    Update $w_t$ over $\mathcal{C}$ using Eq.~\eqref{eq:evidence}\;
    Select $\hat{c}_t$ using Eq.~\eqref{eq:candidate_readout}\;
    Output mesh $\hat{\mathcal{M}}_t=
    (\hat{\mathbf{V}}_t^{\hat{c}_t},\mathcal{T}_{\hat{c}_t})$\;
    \If{$t=T_{\mathrm{id}}$}{
        Retain the $k$ highest-weight candidates in $\mathcal{C}$\;
    }
}
\end{algorithm}

%% file: Images/tikz/fig2_overview.tex
\begin{tikzpicture}[x=0.985mm,y=1mm,font=\footnotesize,
    model/.style={draw=figblue,fill=figblue!6,rounded corners=0.8mm,
        line width=0.65pt,align=center,inner sep=1mm,outer sep=0pt},
    txt/.style={inner sep=0.4pt,outer sep=0pt,align=center,text=black},
    small/.style={txt,font=\scriptsize,text height=1.7mm,text depth=0.5mm},
    bluearrow/.style={-{Latex[length=1.5mm,width=1.15mm]},line width=0.65pt,
        figblue,shorten <=0.4mm,shorten >=0.4mm},
    icon/.style={inner sep=0pt,outer sep=0pt}]

\path[use as bounding box] (0,-5) rectangle (177,58);
\draw[black!18,line width=0.45pt] (75.5,3) -- (75.5,57);
\draw[black!18,line width=0.45pt] (129,3) -- (129,57);
\node[txt,anchor=north,font=\fontsize{8}{9}\selectfont] at (37.75,0)
    {(a) Generative configuration estimation};
\node[txt,anchor=north,font=\fontsize{8}{9}\selectfont] at (102.25,0)
    {(b) Mesh identification};
\node[txt,anchor=north,font=\fontsize{8}{9}\selectfont] at (153,0)
    {(c) Sequential update};
\node[small,anchor=west] at (78,55.5) {Initially: all library meshes};
\node[small,anchor=west] at (132,55.5) {Then: retained top-$k$ meshes};

\node[small] at (11.5,54.5) {Current observation};
\node[icon] (observation-current) at (10,46.5)
    {\includegraphics[height=10mm]{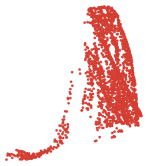}};
\node[model,text width=20mm,minimum height=7mm,font=\scriptsize]
    (encoder) at (37.5,46.5) {Point encoder};

\node[small] at (10,33.5) {Gaussian noise};
\node[icon] (initial) at (10,26.5)
    {\includegraphics[width=11mm,height=11mm,keepaspectratio]{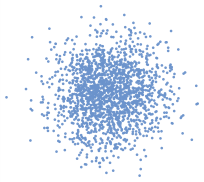}};
\node[model,text width=32.5mm,minimum height=10mm]
    (flow) at (37.5,26.5)
    {\textbf{Graph flow estimator}\\[1.5pt]
     {\scriptsize shared across candidate meshes}};
\node[txt,font=\scriptsize] at (66.5,37.5) {Configuration\\estimate};
\node[icon] (estimate) at (66.5,26.5)
    {\includegraphics[width=14mm,height=16mm,keepaspectratio]{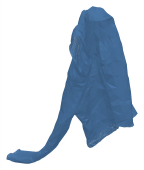}};

\node[icon] (canonical) at (37.5,11.5)
    {\includegraphics[width=14mm,height=9mm,keepaspectratio]{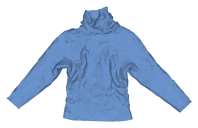}};
\node[small,anchor=north] at (37.5,6) {Candidate canonical mesh};

\draw[bluearrow] (observation-current.east) -- (encoder.west);
\draw[bluearrow] (encoder.south) -- (flow.north);
\draw[bluearrow] (initial.east) -- (flow.west);
\draw[bluearrow] (flow.east) -- (estimate.west);
\draw[bluearrow] (canonical.north) -- (flow.south);

\node[txt,font=\scriptsize] at (85,48.5) {Candidate\\canonical mesh};
\node[txt,font=\scriptsize] at (105,48.5) {Configuration\\estimate};
\node[small] at (121.5,48.5) {Weight};
\draw[figgreen!75!black,rounded corners=0.8mm,line width=0.55pt]
    (78.5,32) rectangle (127,42);
\foreach \y/\k/\w in {37/1/9.4, 25.5/2/1.1, 14/3/0.4}{
    \node[icon] (candidate-\k) at (85,\y)
        {\includegraphics[width=12mm,height=8.5mm,keepaspectratio]{Images/fig2/b_rest_\k.pdf}};
    \node[icon] (configuration-\k) at (105,\y)
        {\includegraphics[width=16mm,height=9mm,keepaspectratio]{Images/fig2/b_recon_\k.pdf}};
    \draw[bluearrow] (91.1,\y) -- (96.9,\y);
    \fill[black!7,rounded corners=0.35mm] (116.5,\y-0.85) rectangle (126.5,\y+0.85);
    \fill[figblue!75!black,rounded corners=0.35mm] (116.5,\y-0.85) rectangle (116.5+\w,\y+0.85);
}
\node[small] at (102.5,5.5) {Estimate $\to$ score $\to$ retain top-$k$};

\node[txt,font=\scriptsize] at (139,48.5) {Retained\\canonical mesh};
\node[txt,font=\scriptsize] at (158.5,48.5) {Current\\estimate};
\node[small] at (172,48.5) {Weight};
\coordinate (update-boundary) at (151.5,26.5);
\draw[figgreen!75!black,rounded corners=0.8mm,line width=0.55pt]
    (update-boundary) rectangle (176.8,39.5);
\foreach \y/\k/\w in {33/1/8.2, 16/2/3.4}{
    \node[small] at (139,\y+7) {Mesh candidate \k};
    \node[icon] (previous-\k) at (139,\y)
        {\includegraphics[width=12mm,height=8mm,keepaspectratio]{Images/fig2/c_rest_\k.pdf}};
    \node[icon] (updated-\k) at (158.5,\y)
        {\includegraphics[width=12mm,height=12mm,keepaspectratio]{Images/fig2/c_recon_\k.pdf}};
    \draw[bluearrow] (previous-\k.east) -- (update-boundary |- updated-\k.west);
    \fill[black!7,rounded corners=0.35mm] (166.3,\y-0.85) rectangle (176.3,\y+0.85);
    \fill[figblue!75!black,rounded corners=0.35mm] (166.3,\y-0.85) rectangle (166.3+\w,\y+0.85);
}
\node[small] at (154.5,5.5) {Accumulate $w_{t-1}$ $\to$ $w_t$};

\end{tikzpicture}

%% file: sections/05_Experiments.tex
\section{Experiments}
\label{sec:experiments}
\label{sec:result}

We designed the experiments to answer the following questions:
\begin{enumerate}
    \item[\textbf{Q1.}] How accurately does the conditional flow-matching model recover complete
    surfaces and vertex positions for unseen garments with different
    mesh sizes and connectivities?
    \item[\textbf{Q2.}] Can observation fit correctly identify a suitable canonical mesh when the true identity is unknown?
    \item[\textbf{Q3.}] Does the estimated mesh correctly localize grasp points on
    real garments after they become occluded?
\end{enumerate}

Q1 evaluates configuration estimation with the correct canonical mesh given
as input. Q2 evaluates mesh selection when its identity is unknown, including
a separate test with the exact mesh removed from the library. Q3 evaluates
grasp-target localization on real garments using an estimated mesh.

\subsection{Experimental Setup}
\label{sec:exp-setup}

\begin{figure}[!b]
    \centering
    \input{Images/tikz/garment_diversity_gallery}
    \caption{Representative canonical meshes from the retrieval library's
    26 dresses, 95 tops, and 100 trousers.}
    \label{fig:library}
\end{figure}
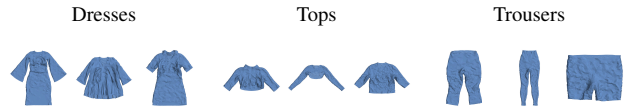

\textbf{Data.}
We collect 1{,}800 simulated manipulation episodes in
DexGarmentLab~\cite{wang2025dexgarmentlab}, covering 221 garments: 26 dresses,
95 tops, and 100 trousers. Each episode begins with a flat garment that one
gripper lifts (\figref[b]{fig:teaser}) before a second gripper grasps and
spreads it. We record the
mesh, gripper states, and a partial point cloud from a fixed depth camera for
each recorded frame. The garment-disjoint split contains 176 training, 23
validation, and 22 test garments. The retrieval library contains all 221
canonical meshes. Library meshes have 1{,}777--2{,}045 vertices and
2{,}294--3{,}894 faces.
\figref{fig:library} shows
nine meshes from this library.

\textbf{Model and training.}
\methodname{} uses \ac{gps} graph transformer layers~\cite{rampasek2022gps}
conditioned on PointNet++ features~\cite{qi2017pointnetpp}.
Each \ac{gps} layer combines message passing along mesh edges with global
vertex self-attention and cross-attention to the point-cloud features.
During training, the point-cloud encoder receives the current cloud
$\mathbf{P}_t$ and an earlier cloud $\mathbf{P}_{t-\Delta}$,
selected uniformly from the preceding five frames.
We sample the noise level $\sigma=\operatorname{sigmoid}(u)$,
$u\sim\mathcal{N}(0,1)$, and use it to condition the network via
\ac{adalnzero}~\cite{peebles2023dit}.
We train for 20 epochs with batch size 32 and AdamW
($\beta_1=0.9$, $\beta_2=0.999$, $\epsilon=10^{-8}$, weight decay $10^{-2}$).
The learning rate warms up for 2{,}000 steps to $2\times10^{-4}$ and then
follows a cosine decay. Gradient norms are clipped to 1.
Training and inference run on an NVIDIA GH200 120\,GB \acs{gpu}.

\textbf{Training augmentations.}
For each cloud, we drop a fraction $d\sim\mathcal{U}(0,0.5)$ of points,
remove a wedge in its \ac{pca} plane with half-angle
$\omega\sim\mathcal{U}(15^\circ,50^\circ)$ with probability 0.5,
and replace a fraction $o\sim\mathcal{U}(0,0.05)$ with uniform outliers
from its bounding box enlarged by 10\% of its extent.
Retained surface points receive isotropic Gaussian noise with a standard
deviation of 3\,mm. We apply a shared scale
$\alpha\sim\mathcal{U}(0.8,1.25)$ and yaw
$\theta\sim\mathcal{U}(-\pi,\pi)$ about gravity to both clouds, the target
configuration, and the canonical mesh. With probability 0.5, we additionally
rotate the canonical mesh alone about its centroid by
$\varphi\sim\mathcal{U}(-\pi,\pi)$ to train heading resolution.
The canonical mesh's scale relative to the target remains fixed.
All augmentations apply only during training.

\textbf{Inference.}
For each candidate–observation pair, we generate $S=4$ configurations
using $J=30$ Euler steps in Eq.~\eqref{eq:flow_euler}, unless otherwise
stated. For the simulated evaluations, both point-cloud encoder inputs
receive the current observation. When accumulating evidence, we update
candidate weights using Eq.~\eqref{eq:evidence} with $\gamma=1$.
When pruning is enabled, we retain $k=5$ candidates after
$T_{\mathrm{id}}=3$ observations.

\subsection{\textbf{Q1}: Configuration Estimation}
\label{sec:exp-state-sim}

We test whether the shared \ac{fm} estimator in \figref[a]{fig:method}
recovers complete configurations of unseen garments. All compared methods
receive the correct canonical mesh and current observation as input,
isolating configuration estimation from mesh identification.

\textbf{Baselines.}
We compare three baselines. \Ac{ra} aligns the flat canonical mesh by
centroid translation and observation-fit yaw search in $5^\circ$ increments.
\Ac{dgnn} predicts a single configuration, using vertex mean squared error
for training and the same inputs and augmentations as \methodname{}.
It uses the same architecture, optimizer, training horizon, and effective
batch size, but a learning rate of $10^{-3}$ and 500 warm-up steps.
\Ac{cr} first completes the surface with a PointNet++ encoder and a
\ac{pcn}-style decoder, then fits the mesh by principal-axis alignment and
non-rigid Chamfer-distance optimization with a validation-tuned Laplacian prior.
UniClothDiff~\cite{tian2025diffusion} is closely related but excluded given it requires one
generative model per mesh, making evaluation over our 221-candidate library
impractical.

\textbf{Inference Protocol and Evaluation Metrics.}
We evaluate 22 held-out garments across 169 trajectories, for a total of
7{,}537 frames. For each frame, \methodname{} generates $S=4$ configurations
from independent Gaussian noise samples and selects the sample with the lowest
observation-fit error in Eq.~\eqref{eq:likelihood}.

We report three geometric metrics. Full-surface Chamfer distance (CD)
averages bidirectional Euclidean nearest-neighbor distances between
2{,}048 points sampled from each of the predicted and ground-truth surfaces.
Occluded CD applies this measure to camera-hidden regions to evaluate
unobserved geometry. Vertex $p_{90}$ is the within-frame 90th percentile
of Euclidean error between corresponding predicted and ground-truth
vertices, capturing larger local errors. For each metric, we report the
median across evaluation frames in mm.

\begin{figure}[t]
    \centering
    \input{Images/tikz/reconstruction_violins}
    \caption{Surface and vertex errors for unseen garments in mm
    (shared logarithmic scale). CD denotes Chamfer distance.
    Circles mark medians, bars the interquartile range, and diamonds the
    frame-level 99th percentile.}
    \label{fig:reconstruction-violins}
\end{figure}
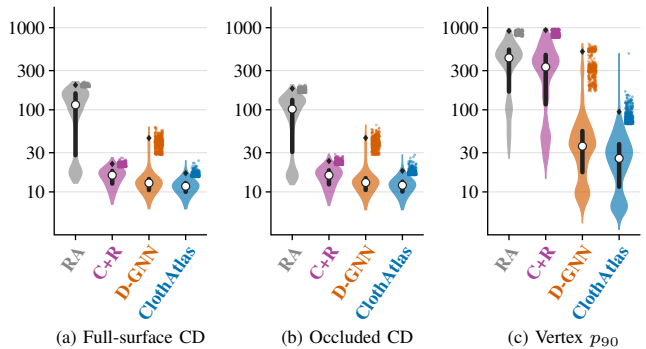

\textbf{Results.}
As shown in \figref{fig:reconstruction-violins}, the shared model achieves
lower median full-surface CD, occluded CD, and vertex $p_{90}$ error than
all three baselines. The first two panels show that its surface
accuracy extends to camera-hidden regions, where the median occluded CD
is 12.0\,mm.
The rightmost panel highlights why surface fit alone is insufficient:
completion followed by registration produces much larger vertex errors
despite a relatively close surface fit. \methodname{} reduces median vertex
$p_{90}$ to 25.6\,mm, compared with 36.1\,mm for the deterministic
network. The shorter upper tails also indicate fewer large errors than the deterministic baseline.

We also test the robustness of \methodname{} to missing points, outliers, and Gaussian noise.
  We evaluate nine observations, every 50 frames from frame 0 to 400,
  from one trajectory per test garment. The configuration columns of
  Table~\ref{tab:robustness} show that \methodname{} retains lower surface
  and vertex errors than the deterministic baseline in every reported
  condition. Missing points affect vertex localization more than surface
  recovery, especially when a contiguous patch is removed. Additional
  Gaussian noise tests with 10\,mm per-coordinate standard deviation
  preserve the surface-accuracy advantage.
\input{Images/data/observation_robustness_compact}

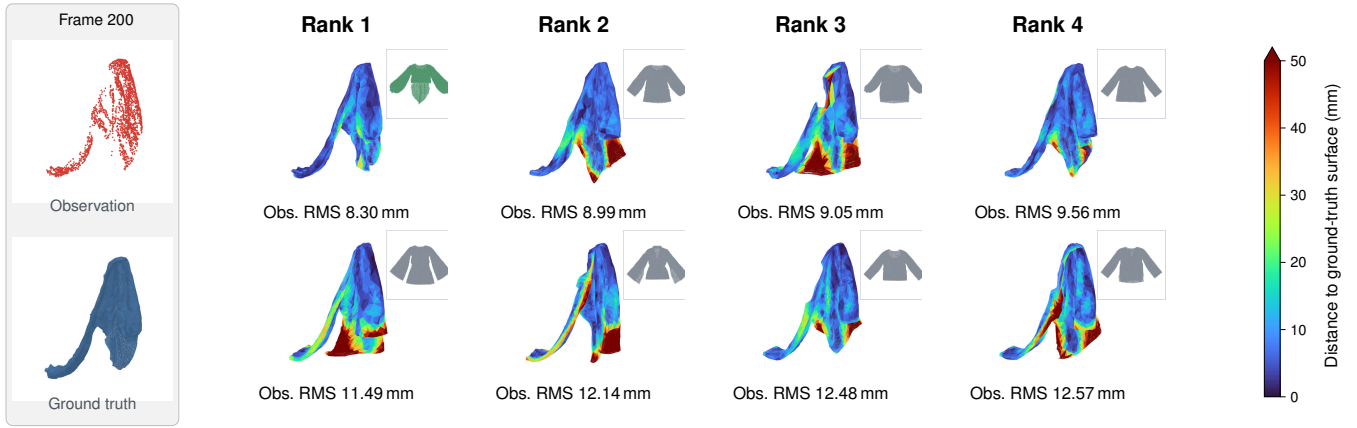
\begin{figure*}[t]
    \centering
    \resizebox{\textwidth}{!}{\input{Images/tikz/reconstruction_ranks.tex}}
    \caption{Top-four candidates for a held-out top at lift frame 200:
    \methodname{} above, \acs{dgnn} below. Observation and ground truth are
    at left while insets show canonical meshes. The correct mesh (green) ranks
    first for \methodname{} and fifth for \acs{dgnn}, outside the displayed
    candidates. Obs. RMS: observation-to-reconstructed-surface root mean
    squared distance in mm.}

    \label{fig:reconstruction-ranks}
\end{figure*}

\subsection{\textbf{Q2}: Canonical-Mesh Identification}
\label{sec:exp-retrieval}
The goal of this evaluation is twofold: (1) to determine whether the proposed ranking procedure, illustrated in \figref[b]{fig:method} and (c), correctly identifies the target mesh from a library of candidates, and (2) to assess whether it retrieves a suitable substitute when the exact mesh is not available.

\textbf{Baselines.}
We compare three baselines: \ac{ra} and \ac{dgnn} from Section~\ref{sec:exp-state-sim}, and \ac{lre}. \ac{ra} ranks aligned meshes by symmetric Chamfer distance, while \ac{dgnn} ranks its predicted configuration for each candidate using the same observation-fit energy as \methodname{}. \ac{lre} instead learns an embedding of partial observations and canonical meshes, and ranks candidates by cosine distance in feature space.

\textbf{Inference Protocol and Evaluation Metrics.}
We evaluate one manipulation episode for each of the 22 test garments,
using nine observations at frames 0, 50, \ldots, 400 for a total of 198
observations. Frame 0 shows the garment lying flat while frame 200 shows it
lifted, illustrated as the leftmost and rightmost meshes in \figref[b]{fig:teaser}.
In the exact-mesh experiment, the library includes the correct mesh for
each test garment, but its identity is not given to the estimator.
These garments are unseen during training, but their canonical meshes
are available at inference. Top-1 and top-5 count how many of the 22
garments have their correct mesh ranked first or among the first five,
respectively, measuring recovery of the reference geometry.

For each observation, \methodname{} scores each candidate by the lowest
observation-fit energy (Eq.~\eqref{eq:likelihood}) among its $S$ configurations.
Table~\ref{tab:identification} compares
independent single-observation rankings with scores accumulated over the
first three observations (frames 0, 50, and 100) or all nine, without
pruning. A separate test retains only the five highest-ranked candidates
after frame 100 and updates them over the remaining observations, as in
\algoref{alg:filter}. Configurations are generated independently for each
observation while candidate scores accumulate across observations.

\begin{table}[t]
    \centering
    \caption{Exact-mesh identification among 221 candidates.
    Cells report top-1 / top-5 counts for 22 test garments
    ($\uparrow$: higher is better). Accumulated scores use $\gamma=1$
    without pruning. Bold marks the best counts.}
    \label{tab:identification}
    \footnotesize
    \setlength{\tabcolsep}{3pt}
    \begin{tabular}{@{}lcccc@{}}
        \toprule
        & \multicolumn{2}{c}{Single observation} & \multicolumn{2}{c}{Accumulated} \\
        \cmidrule(lr){2-3}\cmidrule(l){4-5}
        Method & \shortstack{Lifted\\frame 200} & \shortstack{Flat\\frame 0} & First three & All nine \\
        \midrule
        \acs{ra} & 0 / 1 & 12 / 16 & 11 / 16 & 0 / 1 \\
        \acs{lre} & 1 / 4 & 0 / 5 & 0 / 6 & 1 / 5 \\
        \acs{dgnn} & 8 / 14 & 21 / 21 & 20 / 21 & 13 / 15 \\
        \midrule
        \methodname{} & \textbf{15 / 19} & \textbf{22 / 22} & \textbf{22 / 22} & \textbf{18 / 22} \\
        \bottomrule
    \end{tabular}
\end{table}

\textbf{Results.}
Table~\ref{tab:identification} shows that \methodname{} identifies the
correct mesh more often than all three baselines in each reported setting.
For a lifted garment and using a single observation,
\methodname{} identifies 15 of 22 exact meshes, compared with
8 for \ac{dgnn}. \figref{fig:reconstruction-ranks}
illustrates an example: the correct candidate in green ranks first for
\methodname{} in the upper row, but falls outside the displayed candidates
for \ac{dgnn} below.

The accumulated-score columns of Table~\ref{tab:identification} show that
adding later observations can reverse correct early rankings. With
$\gamma=1$, the update in Eq.~\eqref{eq:evidence} ranks candidates by
their summed observation-fit errors.
Because an incorrect mesh can also fit the observed surface well if the surface is only partially observable, later
score differences can outweigh the correct mesh's early advantage.
Thus, additional observations do not necessarily improve identification.
In the separate pruning test, the five candidates retained after the third
observation include the exact mesh for all 22 garments with \methodname{}
and 21 with \ac{dgnn}. However, the retained meshes can
still be misranked: final top-1 counts are 19 of 22 for \methodname{} and
12 of 22 for \ac{dgnn}.

To test how identification works when we start with a non-flat garment, we replay the manipulation episodes backwards from frame 400 to the flat
frame 0, in intervals of 50 frames.
Starting with uniform weights, we retain five meshes after three observations
with $\gamma=1$. By the final observation, \methodname{} correctly identifies
  16 of 22 garments, compared with 10 of 22 for \ac{dgnn}.
  For both methods, every remaining failure follows pruning of
  the correct mesh, showing that premature candidate removal
  can prevent recovery from later observations.

To test whether a substitute mesh supports surface reconstruction, we remove
each test garment's exact mesh from its candidate library, select a mesh
at frame 0, and fix it over all nine observations. Median full-surface CD
increases from 11.2\,mm with the exact mesh available for selection to
13.4\,mm with it excluded, compared with 14.9\,mm for \ac{dgnn} in the
excluded condition. Both conditions use the same 198 observations.
\methodname{} selects a substitute from the same garment category
  in 19 of 22 cases, compared with 20 of 22 for \ac{dgnn}. Thus, both methods usually recover the garment category, while
  \methodname{} achieves more accurate surface reconstruction.

We also evaluate mesh identification under the same observation corruptions
used for configuration estimation in \secref{sec:exp-state-sim}.
The retrieval columns of Table~\ref{tab:robustness} show that
\methodname{} outperforms \ac{dgnn} after point or patch removal,
but remains sensitive to outliers. Replacing 10\% of observed points
with uniformly sampled points from the observation bounding box
enlarged by 10\% reduces top-1 accuracy from 64.6\% on clean observations
to 2.0\% with $S=4$. Increasing the number of samples does not resolve
this failure. Thus, low configuration errors when the correct mesh
is given do not imply reliable mesh identification. The observation-fit
error in Eq.~\eqref{eq:likelihood} includes every observed point,
making candidate ranking sensitive to segmentation outliers.

Finally, pruning substantially reduces inference cost. Searching all 221 meshes
with $S=4$ and batch size 32 takes 139.1\,s, whereas updating five retained
meshes takes 3.3\,s. %

\subsection{\textbf{Q3}: Robotic Application: Tracking Grasping Points Under Occlusions}
\label{sec:exp-real-grasp}
\label{sec:exp-correspondence}
\label{sec:exp-real-state}
\input{Images/data/real_grasp_tracking}

\begin{figure*}[t]
    \centering
    \input{Images/tikz/real_grasp_blocks}
    \caption{Four separate real-world rollouts compare \methodname{} (upper rows)
    with dense semantic descriptors (DD, lower rows) under temporary occlusion
    (pants, left) and persistent occlusion (top, right). Columns show the initial
    ZED image, initial and lifted representations, and final grasps.
    Green marks G2 and cyan its estimate; orange shows the completed mesh,
    and the heatmap shows descriptor cosine similarity. Checks mark second
    grasps that reached the specified target; crosses mark failures.
    ZED is the sole estimator input; the final RealSense views are used only
    for visualization and annotation.}
    \label{fig:real-grasp}
\end{figure*}
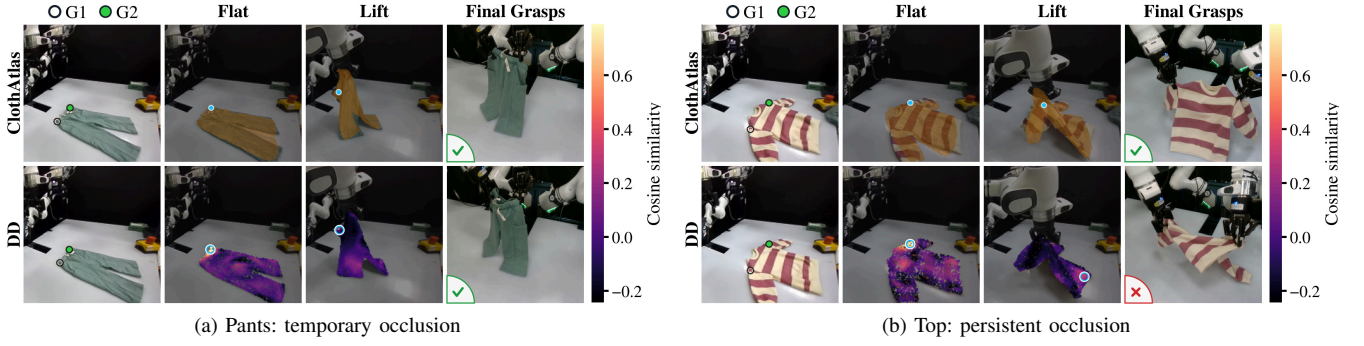

We evaluate \methodname{} on a real-world robotic application: tracking a specified grasp target on garments while the robot lifts them and the target becomes occluded. \figref{fig:real-grasp} illustrates the real-world lifting task under temporary and persistent occlusion.

\textbf{Setup.}
We use the same model as in the previous two sections, trained only in simulation without any real-world data or fine-tuning. We
evaluate \methodname{} on six garments using two Franka Research 3 arms with Robotiq 2F-85 parallel-jaw grippers. First, a user 
specifies two grasp points, G1 and G2, on the initial flat observation. Then, the robot grasps G1 and lifts the garment by 0.30\,m,
displacing G2 and often hiding it from the camera. A calibrated ZED 2 supplies the point cloud segmented with 
the Segment Anything Model (SAM)~\cite{kirillov2023segmentanything}, and a side-mounted RealSense serves as an auxiliary view for manually annotating
occluded grasp points. The only manually selected real-world parameter is one global scale per garment that rescales the real 
garments to simulation dimensions. We leave jointly estimating garment scale and identity to future work.

 \textbf{Tracking Protocol and Evaluation Metrics.}
 We use the inferred mesh configurations to track G2 across observations. G2 is bound to barycentric coordinates on a triangle of the completed mesh and 
evaluates the same coordinates on each subsequent reconstruction. Across 109 trajectories, we manually annotate the final 
position of G2 from the auxiliary views, and call a target occluded when its distance to the nearest point in the final 
mask-filtered ZED cloud exceeds $3$\,cm, giving $62$ visible and $47$ occluded cases. We report the Euclidean error between the
predicted and annotated positions as mean $\pm$ sample standard deviation over the predictions a method returns, and Success@10,
the percentage of trajectories with error below $10$\,cm, counting missing predictions as failures. This tolerance is comparable
 to the gripper's $8.5$\,cm nominal maximum opening.

\textbf{Baselines.}
We compare three baselines. \Ac{sp} predicts that G2 remains at its
  initial position, so its error measures the displacement caused by
  lifting. BootsTAPIR~\cite{doersch2023tapir,doersch2024bootstap}
  tracks G2 in the ZED RGB images; available image predictions are
  converted to \ac{3d} positions using ZED depth.
  \Ac{ddm} uses a dense descriptor model based on
  DexGarmentLab~\cite{wang2025dexgarmentlab} and trained on our simulated
  data. It matches the initial G2 descriptor against descriptors from
  the final masked ZED observation to locate the target via the cosine similarity between the features.
  
\begin{table}[t]
    \centering
    \caption{Grasp-target localization after lifting real garments. Endpoint
    error is mean $\pm$ sample standard deviation in cm ($\downarrow$) over
    available timestamp-aligned predictions. Success@10 is in \% ($\uparrow$),
    with unavailable predictions counted as failures. Vis. and occ. denote the
    62 visible and 47 occluded endpoints under the fixed 3\,cm ZED-support rule.}
    \label{tab:real-grasp}
    \scriptsize
    \setlength{\tabcolsep}{1.5pt}
    \resizebox{\columnwidth}{!}{%
    \begin{tabular}{@{}lcccccc@{}}
        \toprule
        & \multicolumn{3}{c}{Mean $\pm$ SD error $\downarrow$} & \multicolumn{3}{c}{Success@10 $\uparrow$} \\
        \cmidrule(lr){2-4}\cmidrule(l){5-7}
        Method & All & Vis. & Occ. & All & Vis. & Occ. \\
        \midrule
        \acs{sp} & \GCEGraspStaticMeanAll & \GCEGraspStaticMeanVis & \GCEGraspStaticMeanOcc & \GCEGraspStaticSuccAll & \GCEGraspStaticSuccVis & \GCEGraspStaticSuccOcc \\
        BootsTAPIR & \GCEGraspRGBMeanAll & \GCEGraspRGBMeanVis & \GCEGraspRGBMeanOcc & \GCEGraspRGBSuccAll & \GCEGraspRGBSuccVis & \GCEGraspRGBSuccOcc \\
        \acs{ddm} & \GCEGraspDescMeanAll & \GCEGraspDescMeanVis & \GCEGraspDescMeanOcc & \GCEGraspDescSuccAll & \GCEGraspDescSuccVis & \GCEGraspDescSuccOcc \\
        \methodname{} & \textbf{\GCEGraspOursMeanAll} & \textbf{\GCEGraspOursMeanVis} & \textbf{\GCEGraspOursMeanOcc} & \textbf{\GCEGraspOursSuccAll} & \textbf{\GCEGraspOursSuccVis} & \textbf{\GCEGraspOursSuccOcc} \\
        \bottomrule
    \end{tabular}}
\end{table}

 \textbf{Results.} Table~\ref{tab:real-grasp} shows that binding the target to the estimated
 mesh localizes it more accurately than descriptor matching or image
 tracking, and the margin is largest when the grasping point is occluded.
 On the 47 occluded endpoints, \methodname{} reaches 85.1\% Success@10
 against 36.2\% for \ac{ddm}, and it returns a prediction for 46 of them,
 whereas BootsTAPIR returns one for only 10: a 2D track without visible depth
 yields no graspable position. %
 
\figref{fig:real-grasp} shows qualitative results for \methodname{} and \ac{ddm} under two occlusion cases: self-occlusion with and without re-entering the ZED view. Unlike descriptors, which can only point at what
the camera currently observes, a cue defined on the mesh stays addressable
regardless of visibility. %

\textbf{Failure modes.}
\methodname{} fails when no library candidate is geometrically close enough to
the real garment, so the estimator cannot deform it to fit the observed point
cloud. Separately, the simulated prior models garments as thin shells without
the volume of stacked fabric layers, so real deformation in hidden regions can
depart from what the model predicts and mislocalize an occluded target.

%% file: Images/tikz/garment_diversity_gallery.tex
\begingroup%
\ifdefined\garmentgalleryunit\else\newlength{\garmentgalleryunit}\fi%
\setlength{\garmentgalleryunit}{\dimexpr\columnwidth*2/7\relax}%
\input{Images/fig1/gallery_metadata.tex}%
\begin{tikzpicture}[x=\garmentgalleryunit,y=-\garmentgalleryunit,
    label/.style={anchor=north,inner sep=0pt,outer sep=0pt,
        font=\fontsize{7.8}{9}\selectfont},
    icon/.style={anchor=north west,inner sep=0pt,outer sep=0pt}]
\path[use as bounding box] (0,0) rectangle (3.5,.77);
\foreach \category/\categorylabel/\total/\first/\second/\third [count=\group from 0]
    in \figonegalleryrows {
    \pgfmathsetmacro{\groupx}{.10+\group*1.14}
    \node[label] at (\groupx+.51,.04) {\categorylabel};
    \foreach \asset [count=\position from 0] in {\first,\second,\third}{
        \node[icon] at (\groupx+\position*.35,.26)
            {\includegraphics[width=.32\garmentgalleryunit]{Images/fig4/\category_\asset.pdf}};
    }
}
\end{tikzpicture}%
\endgroup%

%% file: Images/tikz/reconstruction_violins.tex
\begingroup%
\input{Images/tikz/reconstruction_violin_style.tex}%
\def\cpviolinmethodlabels{0/RA/rigid,1/C+R/completion,2/D-GNN/deterministic,3/\methodname{}/independent}%
\def\cpviolinlabelanchor{north east}%
\def\cpviolinlabelangle{55}%
\pgfmathsetlengthmacro{\cpviolinaxiswidth}{.234*\linewidth}%
\begin{tikzpicture}[x=\linewidth,y=1in,
    font=\fontsize{7}{8.4}\selectfont,
    cp label/.style={anchor=base west,inner sep=0pt,outer sep=0pt}]
\node[anchor=base,inner sep=0pt] at (.212,.06) {(a) Full-surface \acs{cd}};
\node[anchor=base,inner sep=0pt] at (.544,.06) {(b) Occluded \acs{cd}};
\node[anchor=base,inner sep=0pt] at (.876,.06) {(c) Vertex $p_{90}$};

    \begin{scope}[shift={(.095\linewidth,.625in)},x=\cpviolinaxiswidth,y=1.188in]
    \pgfmathsetmacro{\cplogmin}{ln(\cpviolinmin)/ln(10)}
    \pgfmathsetmacro{\cplogmax}{ln(\cpviolinmax)/ln(10)}
    \pgfmathsetmacro{\cplogspan}{\cplogmax-\cplogmin}
    \edef\cpticklist{\cpviolinticks}
    \foreach \tick in \cpticklist {
        \pgfmathsetmacro{\tickpos}{(ln(\tick)/ln(10)-\cplogmin)/\cplogspan}
        \draw[black!13,line width=.35pt] (0,\tickpos) -- (1,\tickpos);
        \draw[black,line width=.5pt] (0,\tickpos) -- ++(-2pt,0);
        \node[anchor=east,inner sep=0pt,outer sep=0pt]
            at ([xshift=-3.5pt]0,\tickpos) {\tick};
    }
    \pgfmathsetlengthmacro{\cpdatax}{\cpviolinaxiswidth/4.16}
    \pgfmathsetlengthmacro{\cpdatay}{1.188in/\cplogspan}
    \begin{scope}[x=\cpdatax,y=\cpdatay]
        \begin{scope}[shift={(.58,-\cplogmin)}]
            \clip (-.58,\cplogmin) rectangle (3.58,\cplogmax);
            \input{Images/data/reconstruction_violins/chamfer_mm.tex}
        \end{scope}
    \end{scope}
    \draw[black,line width=.5pt] (0,1) -- (0,0) -- (1,0);
    \foreach \methodindex/\methodletter/\methodcolor in \cpviolinmethodlabels {
        \pgfmathsetmacro{\tickpos}{(\methodindex+.58)/4.16}
        \draw[black,line width=.5pt] (\tickpos,0) -- ++(0,-2pt);
        \node[anchor=\cpviolinlabelanchor,rotate=\cpviolinlabelangle,inner sep=0pt,outer sep=0pt,text=cpviolin\methodcolor,font=\bfseries\fontsize{7}{8.4}\selectfont]
            at ([yshift=-3.5pt]\tickpos,0) {\methodletter};
    }
    \end{scope}%

    \begin{scope}[shift={(.427\linewidth,.625in)},x=\cpviolinaxiswidth,y=1.188in]
    \pgfmathsetmacro{\cplogmin}{ln(\cpviolinmin)/ln(10)}
    \pgfmathsetmacro{\cplogmax}{ln(\cpviolinmax)/ln(10)}
    \pgfmathsetmacro{\cplogspan}{\cplogmax-\cplogmin}
    \edef\cpticklist{\cpviolinticks}
    \foreach \tick in \cpticklist {
        \pgfmathsetmacro{\tickpos}{(ln(\tick)/ln(10)-\cplogmin)/\cplogspan}
        \draw[black!13,line width=.35pt] (0,\tickpos) -- (1,\tickpos);
        \draw[black,line width=.5pt] (0,\tickpos) -- ++(-2pt,0);
        \node[anchor=east,inner sep=0pt,outer sep=0pt]
            at ([xshift=-3.5pt]0,\tickpos) {\tick};
    }
    \pgfmathsetlengthmacro{\cpdatax}{\cpviolinaxiswidth/4.16}
    \pgfmathsetlengthmacro{\cpdatay}{1.188in/\cplogspan}
    \begin{scope}[x=\cpdatax,y=\cpdatay]
        \begin{scope}[shift={(.58,-\cplogmin)}]
            \clip (-.58,\cplogmin) rectangle (3.58,\cplogmax);
            \input{Images/data/reconstruction_violins/chamfer_occ_mm.tex}
        \end{scope}
    \end{scope}
    \draw[black,line width=.5pt] (0,1) -- (0,0) -- (1,0);
    \foreach \methodindex/\methodletter/\methodcolor in \cpviolinmethodlabels {
        \pgfmathsetmacro{\tickpos}{(\methodindex+.58)/4.16}
        \draw[black,line width=.5pt] (\tickpos,0) -- ++(0,-2pt);
        \node[anchor=\cpviolinlabelanchor,rotate=\cpviolinlabelangle,inner sep=0pt,outer sep=0pt,text=cpviolin\methodcolor,font=\bfseries\fontsize{7}{8.4}\selectfont]
            at ([yshift=-3.5pt]\tickpos,0) {\methodletter};
    }
    \end{scope}%

    \begin{scope}[shift={(.759\linewidth,.625in)},x=\cpviolinaxiswidth,y=1.188in]
    \pgfmathsetmacro{\cplogmin}{ln(\cpviolinmin)/ln(10)}
    \pgfmathsetmacro{\cplogmax}{ln(\cpviolinmax)/ln(10)}
    \pgfmathsetmacro{\cplogspan}{\cplogmax-\cplogmin}
    \edef\cpticklist{\cpviolinticks}
    \foreach \tick in \cpticklist {
        \pgfmathsetmacro{\tickpos}{(ln(\tick)/ln(10)-\cplogmin)/\cplogspan}
        \draw[black!13,line width=.35pt] (0,\tickpos) -- (1,\tickpos);
        \draw[black,line width=.5pt] (0,\tickpos) -- ++(-2pt,0);
        \node[anchor=east,inner sep=0pt,outer sep=0pt]
            at ([xshift=-3.5pt]0,\tickpos) {\tick};
    }
    \pgfmathsetlengthmacro{\cpdatax}{\cpviolinaxiswidth/4.16}
    \pgfmathsetlengthmacro{\cpdatay}{1.188in/\cplogspan}
    \begin{scope}[x=\cpdatax,y=\cpdatay]
        \begin{scope}[shift={(.58,-\cplogmin)}]
            \clip (-.58,\cplogmin) rectangle (3.58,\cplogmax);
            \input{Images/data/reconstruction_violins/vert_err_p90_mm.tex}
        \end{scope}
    \end{scope}
    \draw[black,line width=.5pt] (0,1) -- (0,0) -- (1,0);
    \foreach \methodindex/\methodletter/\methodcolor in \cpviolinmethodlabels {
        \pgfmathsetmacro{\tickpos}{(\methodindex+.58)/4.16}
        \draw[black,line width=.5pt] (\tickpos,0) -- ++(0,-2pt);
        \node[anchor=\cpviolinlabelanchor,rotate=\cpviolinlabelangle,inner sep=0pt,outer sep=0pt,text=cpviolin\methodcolor,font=\bfseries\fontsize{7}{8.4}\selectfont]
            at ([yshift=-3.5pt]\tickpos,0) {\methodletter};
    }
    \end{scope}%

\pgfresetboundingbox
\path[use as bounding box] (0,0) rectangle (1,1.84);
\end{tikzpicture}%
\endgroup%

%% file: Images/data/observation_robustness_compact.tex
\begin{table}[t]
\centering
\caption{Configuration estimation and exact-mesh retrieval under corrupted
observations. Surface Chamfer distance (CD) and vertex $p_{90}$ are median
errors in mm ($\downarrow$), with the correct mesh supplied. Retrieval
reports top-1 / top-5 counts ($\uparrow$), ranking all 221 meshes separately
per observation. Counts are out of 198 for clean observations and 594 for
each corruption (three realizations per observation).
Bold marks the best values within each condition. Ours: \methodname{}.}
\label{tab:robustness}
\footnotesize
\setlength{\tabcolsep}{2.5pt}
\begin{tabular}{@{}llrrc@{}}
\toprule
 & & \multicolumn{2}{c}{Configuration} & Retrieval \\
\cmidrule(lr){3-4}\cmidrule(l){5-5}
Condition & Method & CD & Vertex $p_{90}$ & Top-1 / Top-5 \\
\midrule
Clean & \acs{dgnn} & 12.28 & 35.5 & 92 / 130 \\
 & Ours, $S=4$ & \textbf{11.21} & \textbf{24.8} & \textbf{128 / 170} \\
\midrule
90\% removal & \acs{dgnn} & 15.36 & 66.8 & 118 / 246 \\
 & Ours, $S=1$ & 12.92 & 42.8 & 254 / 394 \\
 & Ours, $S=4$ & 12.60 & 37.7 & 285 / 424 \\
 & Ours, $S=8$ & \textbf{12.49} & \textbf{36.6} & \textbf{301 / 442} \\
\midrule
50\% patch & \acs{dgnn} & 15.72 & 59.2 & 159 / 260 \\
 & Ours, $S=4$ & \textbf{14.26} & \textbf{47.8} & \textbf{275 / 374} \\
\midrule
10\% outliers & \acs{dgnn} & 13.39 & 40.3 & 5 / 45 \\
 & Ours, $S=4$ & \textbf{12.37} & \textbf{34.8} & \textbf{12 / 47} \\
\bottomrule
\end{tabular}
\end{table}

%% file: Images/tikz/reconstruction_ranks.tex
\begingroup%
\input{Images/reconstruction_ranks/values.tex}%
\definecolor{rrReferenceText}{HTML}{4B535C}%
\begin{tikzpicture}[x=1mm,y=1mm,
    font=\sffamily\fontsize{7}{8.4}\selectfont,
    label/.style={inner sep=0pt,outer sep=0pt,anchor=base,align=center},
    panel/.style={inner sep=0pt,outer sep=0pt,anchor=south west},
    rank/.style={label,font=\sffamily\bfseries\fontsize{9}{10.8}\selectfont},
    model/.style={inner sep=0pt,outer sep=0pt,rotate=90,
        font=\sffamily\bfseries\fontsize{8}{9.6}\selectfont}]

\path[use as bounding box] (0,12) rectangle (196,74.93);
\filldraw[fill=black!5,draw=black!25,line width=0.4pt,rounded corners=1mm]
    (0,12.8) rectangle (25.2,74.3);
\node[label,font=\sffamily\fontsize{6.5}{7.8}\selectfont]
    at (13.1674,71.2584) {Frame 200};

\def\rrFirstColumn{29.9184}
\def\rrColumnPitch{34.576}
\def\rrPanelWidth{36.576}
\def\rrPanelHeight{23.0035}
\def\rrInsetSize{10}
\foreach \r in {1,2,3,4}{
    \pgfmathsetmacro{\cx}{\rrFirstColumn+(\r-1)*\rrColumnPitch+\rrPanelWidth/2}
    \node[rank] at (\cx,70.2584) {Rank \r};
}

\node[panel] at (0.9144,45.3327)
    {\includegraphics[width=23.4086mm,height=23.6779mm]{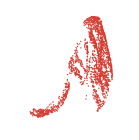}};
\node[panel] at (0.9144,16.4846)
    {\includegraphics[width=23.4086mm,height=23.6779mm]{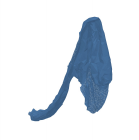}};
\node[label,text=rrReferenceText,font=\sffamily\fontsize{6.5}{7.8}\selectfont]
    at (12.6187,44.0588) {Observation};
\node[label,text=rrReferenceText,font=\sffamily\fontsize{6.5}{7.8}\selectfont]
    at (12.6187,15.2108) {Ground truth};

\foreach \m/\name/\base in {flow/\methodname{}/42.8341,deterministic/\acs{dgnn}/16.4860}{
    \pgfmathsetmacro{\meshY}{\base+2.7724}
    \pgfmathsetmacro{\scoreY}{\base+0.2248}
    \pgfmathsetmacro{\modelY}{\base+12.9629}
    \node[model] at (33.1635,\modelY) {\name};
    \foreach \r in {1,2,3,4}{
        \pgfmathsetmacro{\px}{\rrFirstColumn+(\r-1)*\rrColumnPitch}
        \pgfmathsetmacro{\cx}{\px+\rrPanelWidth/2}
        \node[panel] at (\px,\meshY)
            {\includegraphics[width=\rrPanelWidth mm,height=\rrPanelHeight mm]
                {Images/reconstruction_ranks/\m_rank\r.pdf}};
        \node[panel] at ({\px+25.5},{\meshY+12})
            {\includegraphics[width=\rrInsetSize mm]
                {Images/reconstruction_ranks/\m_rank\r_inset.pdf}};
        \node[label] at (\cx,\scoreY)
            {Obs. \acs{rms} \csname rrScore\m\r\endcsname\,mm};
    }
}

\node[inner sep=0pt,rotate=90] at (193.5,42)
    {Distance to ground-truth surface (mm)};
\begin{scope}[shift={(183.5,17)}]
    \node[panel,rotate=90] at (2.3,0)
        {\resizebox{49mm}{2.3mm}{\pgfuseshading{rrErrorGradient}}};
    \fill[rrError255] (0,49) -- (1.15,51) -- (2.3,49) -- cycle;
    \draw[line width=0.4pt] (0,0) -- (2.3,0) -- (2.3,49)
        -- (1.15,51) -- (0,49) -- cycle;
    \foreach \tick in {0,10,...,50}{
        \draw[line width=0.4pt] (2.3,{49*\tick/50}) -- (3.1,{49*\tick/50});
        \node[inner sep=0pt,outer sep=0pt,anchor=west,
            font=\sffamily\fontsize{6}{7.2}\selectfont]
            at (3.8,{49*\tick/50}) {\tick};
    }
\end{scope}
\end{tikzpicture}%
\endgroup%

%% file: Images/data/real_grasp_tracking.tex
\newcommand{\GCEGraspStaticMeanAll}{\(18.41 \mathbin{\pm} 7.54\)}
\newcommand{\GCEGraspStaticMeanVis}{\(18.55 \mathbin{\pm} 7.03\)}
\newcommand{\GCEGraspStaticMeanOcc}{\(18.22 \mathbin{\pm} 8.24\)}
\newcommand{\GCEGraspStaticSuccAll}{2.8}
\newcommand{\GCEGraspStaticSuccVis}{1.6}
\newcommand{\GCEGraspStaticSuccOcc}{4.3}
\newcommand{\GCEGraspRGBMeanAll}{\(15.75 \mathbin{\pm} 8.25\)}
\newcommand{\GCEGraspRGBMeanVis}{\(15.93 \mathbin{\pm} 7.58\)}
\newcommand{\GCEGraspRGBMeanOcc}{\(15.64 \mathbin{\pm} 9.02\)}
\newcommand{\GCEGraspRGBSuccAll}{1.8}
\newcommand{\GCEGraspRGBSuccVis}{0.0}
\newcommand{\GCEGraspRGBSuccOcc}{4.3}
\newcommand{\GCEGraspDescMeanAll}{\(15.44 \mathbin{\pm} 10.12\)}
\newcommand{\GCEGraspDescMeanVis}{\(14.27 \mathbin{\pm} 10.05\)}
\newcommand{\GCEGraspDescMeanOcc}{\(16.99 \mathbin{\pm} 10.11\)}
\newcommand{\GCEGraspDescSuccAll}{34.9}
\newcommand{\GCEGraspDescSuccVis}{33.9}
\newcommand{\GCEGraspDescSuccOcc}{36.2}
\newcommand{\GCEGraspOursMeanAll}{\(6.38 \mathbin{\pm} 6.71\)}
\newcommand{\GCEGraspOursMeanVis}{\(4.91 \mathbin{\pm} 3.17\)}
\newcommand{\GCEGraspOursMeanOcc}{\(8.33 \mathbin{\pm} 9.27\)}
\newcommand{\GCEGraspOursSuccAll}{91.7}
\newcommand{\GCEGraspOursSuccVis}{96.8}
\newcommand{\GCEGraspOursSuccOcc}{85.1}
\newcommand{\GCEGraspDescSuccFiveAll}{12.8}
\newcommand{\GCEGraspDescSuccFiveOcc}{2.1}
\newcommand{\GCEGraspOursSuccFiveAll}{52.3}
\newcommand{\GCEGraspOursSuccFiveOcc}{36.2}

%% file: Images/tikz/real_grasp_blocks.tex
\subfloat[Pants: temporary occlusion]{%
    \includegraphics[width=0.495\textwidth]{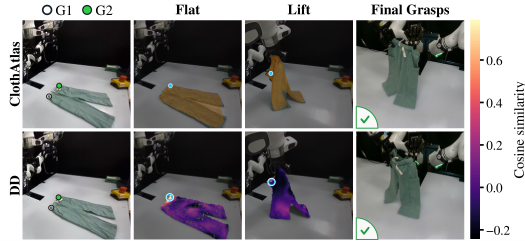}}%
\hfill
\subfloat[Top: persistent occlusion]{%
    \includegraphics[width=0.495\textwidth]{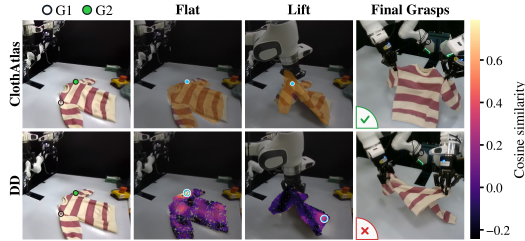}}

%% file: sections/06_Conclusion.tex
\section{Conclusion}
\label{sec:conclusion}
We presented \methodname{}, which reconstructs complete garment
  configurations from partial point clouds with a shared graph-conditioned
  \ac{fm} model and identifies canonical meshes by accumulated observation
  fit. Trained entirely in simulation without action supervision, it accepts
  unseen canonical meshes without retraining and preserves vertex identities
  for localizing targets through deformation and occlusion. From a single
  self-occluded observation, it retrieves 15/22 exact meshes among 221
  candidates, versus 8/22 for a deterministic graph estimator. On real
  garments, it localizes 85.1\% of occluded grasp targets within 10\,cm,
  versus 36.2\% for dense descriptors.
  
  Future work should
  investigate faster, more robust candidate evaluation and active
  perception to resolve ambiguous geometry. More broadly, a complete mesh with persistent vertex identities is a task-agnostic interface to garment state. It supports specifying where to act on a garment, predicting the effect of candidate actions with mesh-based dynamics models, and verifying goal conditions that depend on hidden fabric layers. Connecting this interface to downstream decision making, including generalist policies and high-level planners, is a natural next step.